\documentclass{article}
\usepackage{times}
\usepackage[hyphens]{url}
\usepackage{hyperref}
\usepackage{graphicx}
\usepackage{subfig}
\usepackage{amsmath}
\usepackage{amsthm}
\usepackage{amsfonts}
\usepackage{amssymb}
\usepackage{xcolor}
\usepackage{enumitem}
\usepackage{caption}
\usepackage{marginnote}
\usepackage{dialogue}
\usepackage{fancyvrb}
\usepackage[accepted]{icml2019}
\usepackage{natbib}
\usepackage{verbatim}

\begin{document}

\twocolumn[
\icmltitle{
CraftAssist: A Framework for Dialogue-enabled Interactive Agents 
}

\icmlsetsymbol{equal}{*}
\begin{icmlauthorlist}
\icmlauthor{Jonathan Gray}{equal}
\icmlauthor{Kavya Srinet}{equal}
\icmlauthor{Yacine Jernite}{}
\icmlauthor{Haonan Yu}{}
\icmlauthor{Zhuoyuan Chen}{}
\icmlauthor{Demi Guo}{}
\icmlauthor{Siddharth Goyal}{}
\icmlauthor{C. Lawrence Zitnick}{}
\icmlauthor{Arthur Szlam}{}
\vskip 0.1in
Facebook AI Research
\vskip 0.1in
\small{\texttt{\{jsgray,ksrinet\}@fb.com}}
\vskip 0.35in
\end{icmlauthorlist}

%\icmlaffiliation{to}{Department of Computation, University of Torontoland, Torontoland, Canada}
%\icmlcorrespondingauthor{Jonathan Gray}{jsgray@fb.com}

]
{\let\thefootnote\relax\footnotetext{\hspace*{-\footnotesep}\textsuperscript{*}   Equal contribution
}}
% % otherwise use the standard text.

%%%%%%%%%%%%%%%%%%%%%%%%%%%%%%%%%%%%%%%%%%%%%%%%%%%%%%%%%%%%%%%%%%%%%%%%%%%%%%%%

\begin{abstract}
This paper describes an implementation of a bot assistant in Minecraft, and the tools and platform allowing players to interact with the bot and to record those interactions.    The purpose of building such an assistant is to facilitate the study of agents that can complete tasks specified by dialogue, and eventually, to learn from dialogue interactions.
%We propose an environment for studying bots that interact with humans in open-ended domains via natural language. We work in the sandbox construction game of Minecraft that enables a bot to interact with human players in a multiplayer setting. We describe an initial implementation of a bot whose goal is to be a useful assistant to humans. The code, data, and models are open-sourced to the community to help further research in this area. 
\end{abstract}

%%%%%%%%%%%%%%%%%%%%%%%%%%%%%%%%%%%%%%%%%%%%%%%%%%%%%%%%%%%%%%%%%%%%%%%%%%%%%%%%
\section{Introduction}
While machine learning (ML) methods have achieved impressive performance on difficult but narrowly-defined tasks \cite{alphago,he2017mask, mahajan2018exploring, atari}, building more general systems that perform well at a variety of tasks remains an area of active research. Here we are interested in systems that are competent in a long-tailed distribution of simpler tasks, specified (perhaps ambiguously) by humans using natural language. As described in our position paper \cite{positionPaper}, we propose to study such systems through the development of an assistant bot in the open sandbox game of Minecraft\footnote{Minecraft features: \textcopyright Mojang Synergies AB included courtesy of Mojang AB}~\cite{johnson2016malmo,MineRL}. This paper describes the implementation of such a bot, and the tools and platform allowing players to interact with the bot and to record those interactions.

\begin{figure}
    \centering \includegraphics[width=0.45\textwidth]{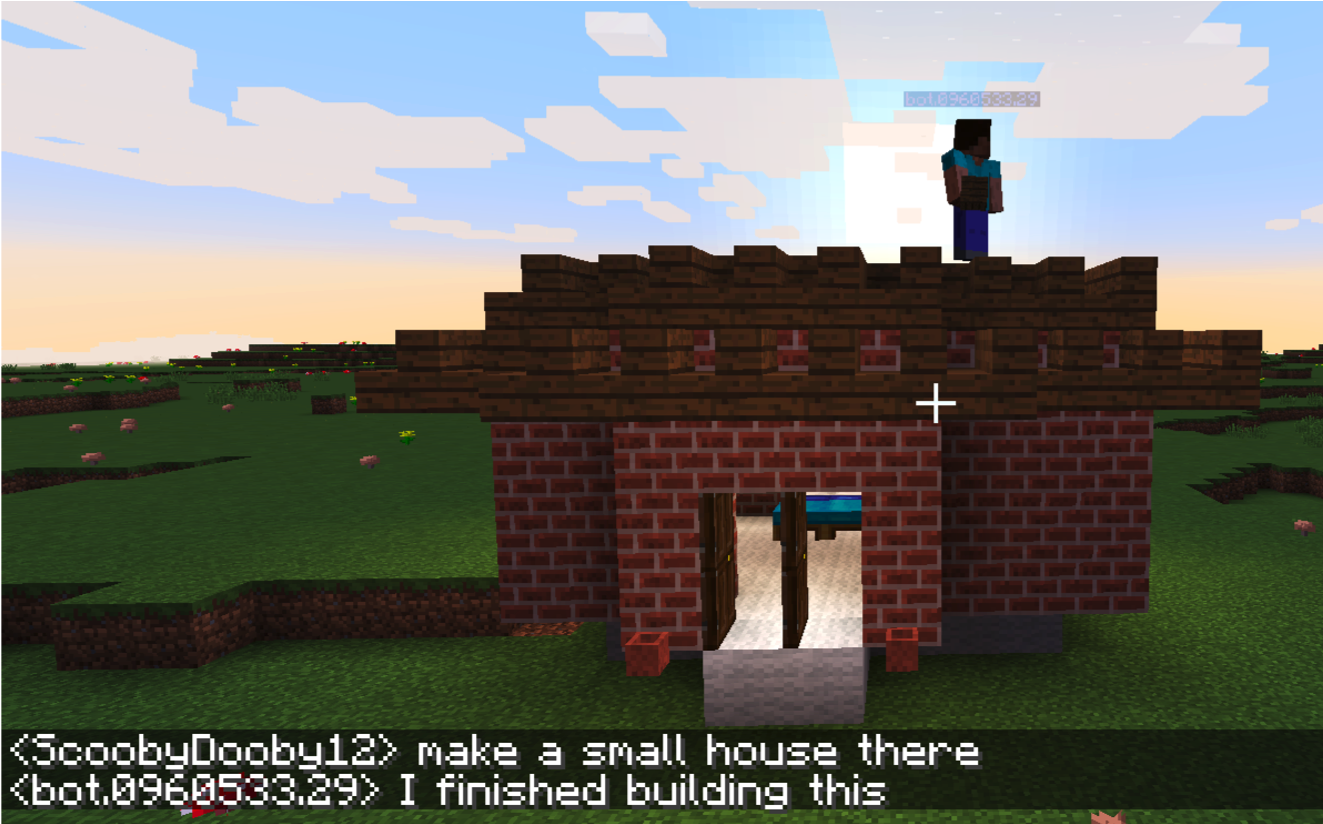}
    \caption{An in-game screenshot of a human player using in-game chat to communicate with the bot.}
    \label{fig:screenshot_bot}
\end{figure}

The bot appears and interacts like another player:  other players can observe the bot moving around and modifying the world, and communicate with it via in-game chat. Figure \ref{fig:screenshot_bot} shows a screenshot of a typical in-game experience. Neither Minecraft nor the software framework described here provides an explicit objective or reward function; 
the ultimate goal of the bot is to be a useful and fun assistant in a wide variety of tasks specified and evaluated by human players.  

Longer term, we hope to build assistants that interact and collaborate with humans to actively learn new concepts and skills. However, the bot described here should be taken as initial point from which we (and others) can iterate.  As the bots become more capable, we can expand the scenarios where they can effectively learn. 

To encourage collaborative research, the code, data, and models are open-sourced\footnote{\url{https://github.com/facebookresearch/craftassist}}. The design of the framework is purposefully modular to allow research on components of the bot as well as the whole system. The released data includes the human actions used to build 2,586 houses, the labeling of the sub-parts of the houses (e.g., walls, roofs, etc.), human rewordings of templated commands, and the mapping of natural language commands to bot interpretable logical forms. To enable researchers to independently collect data, the infrastructure that allows for the recording of human and bot interaction on a Minecraft server is also released. We hope these tools will help empower research on agents that can complete tasks specified by dialogue, and eventually, learn form dialogue interactions.

%\begin{itemize}
%    \item What interaction with the Assistant looks like: the bot is another player, communication via in-game chat
%    \item Project goals: no intrinsic environment reward, reference position paper
%    \item Example dialogues
%\end{itemize}

\section{Minecraft}
Minecraft\footnote{\url{https://minecraft.net/en-us/}} is a popular multiplayer open world voxel-based building and crafting game. Gameplay starts with a procedurally generated world containing natural features (e.g. trees, mountains, and fields) all created from an atomic set of a few hundred possible blocks. Additionally, the world is populated from an atomic set of animals and other non-player characters, commonly referred to as ``mobs''.

\begin{figure}
    \centering \includegraphics[width=0.45\textwidth]{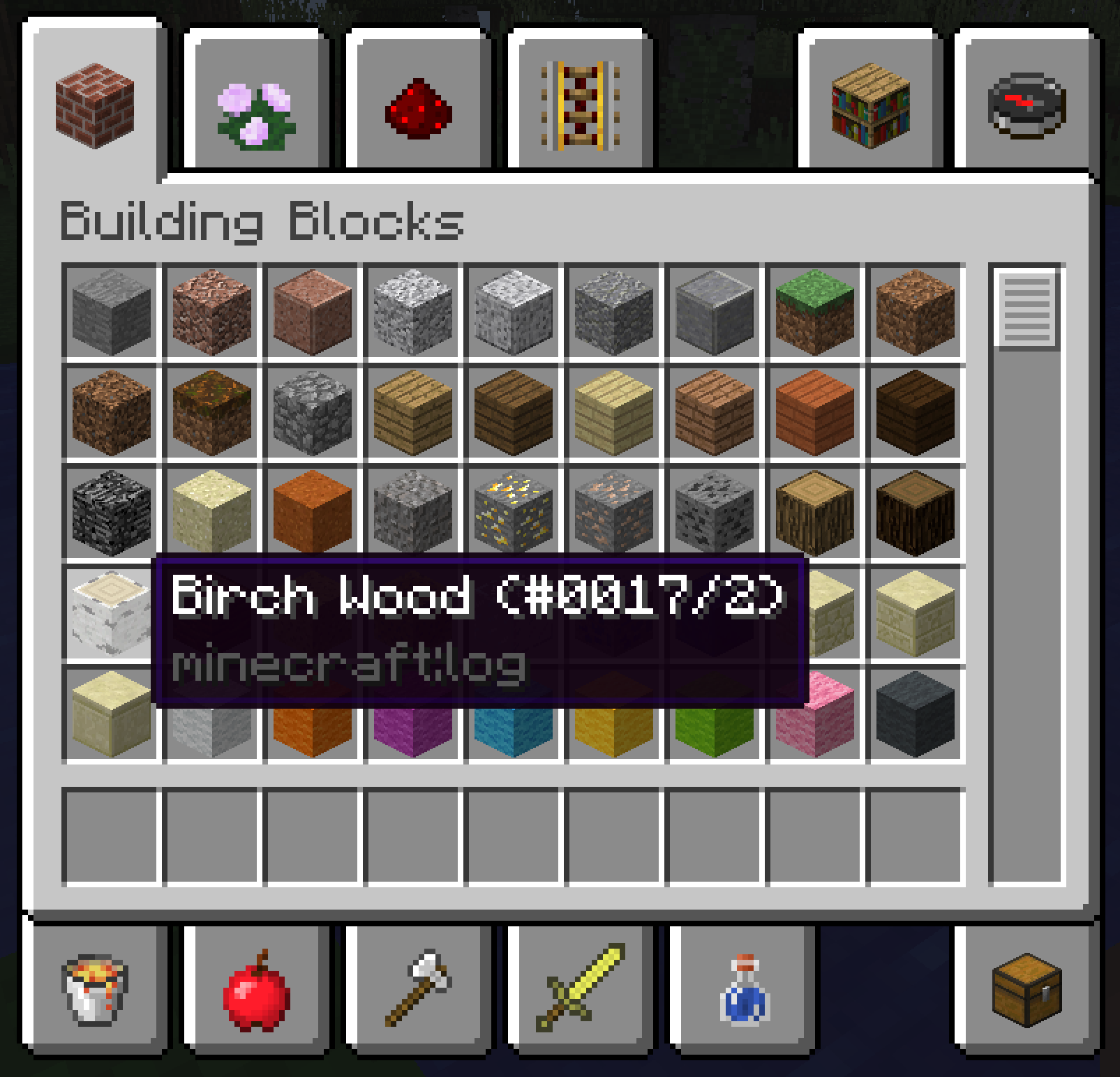}
    \caption{An in-game screenshot showing some of the block types available to the user in creative mode.}
    \label{fig:screenshot_blocks}
\end{figure}

The game has two main modes: ``creative'' and ``survival''. In survival mode the player is resource limited, can be harmed, and is subject to more restrictive physics.  In creative mode, the player is not resource limited, cannot be harmed, and is subject to less restrictive physics, e.g. the player can fly through the air. An in-depth guide to Minecraft can be found at \url{https://minecraft.gamepedia.com/Minecraft}.

%\lz{Players may build objects of two types; atomic objects and compound objects.} \jgray{This terminology is a little confusing. I'd rather stay in terms of blocks, block types, and the voxel grid. TODO} Atomic objects are created through a process called ``crafting'' which requires the player to gather and combine multiple atomic objects to create a new atomic object, such as combining three wood blocks and three wool to create a bed. 
In survival mode, blocks can be combined in a process called ``crafting'' to create other blocks. For example, three wood blocks and three wool can be combined to create an atomic ``bed'' block. In creative mode, players have access to all block types without the need for crafting.

Compound objects are arrangements of multiple atomic objects, such as a house constructed from brick, glass and door blocks. Players may build compound objects in the world by placing or removing blocks of different types in the environment. Figure \ref{fig:screenshot_blocks} shows a sample of different block types. The blocks are placed on a 3D voxel grid. Each voxel in the grid contains one material. In this paper, we assume players are in creative mode and we focus on building compound objects.

Minecraft, particularly in its creative mode setting, has no win condition and encourages players to be creative. The diversity of objects created in Minecraft is astounding; these include landmarks, sculptures, temples, roller-coasters and entire cityscapes. Collaborative building is a common activity in Minecraft.

Minecraft allows multiplayer servers, and players can collaborate to build, survive, or compete. It has a huge player base (91M monthly active users in October 2018) \footnote{\url{https://www.gamesindustry.biz/articles/2018-10-02-minecraft-exceeds-90-million-monthly-active-users}}, and players actively create game mods and shareable content.  The multiplayer game has a built-in text chat for player to player communication. Dialogue between users on multi-user servers is a standard part of the game.

\section{Client/Server Architecture}
Minecraft operates through a client and server architecture. The bot acting as a client communicates with the Minecraft server using the Minecraft network protocol\footnote{We have implemented protocol version 340, which corresponds to Minecraft Computer Edition v1.12.2, and is described here: \href{https://wiki.vg/index.php?title=Protocol\&oldid=14204}{https://wiki.vg/index.php?title=Protocol\&oldid=14204}}. The server may receive actions from multiple bot or human clients, and returns world updates based on player and mob actions. Our implementation of a Minecraft network client is included in the top-level \href{https://github.com/facebookresearch/craftassist/tree/master/client}{client} directory.

Implementing the Minecraft protocol enables a bot to connect to any Minecraft server without the need for installing server-side mods, when using this framework. This provides two main benefits:

\begin{enumerate}
    \item A bot can easily join a multiplayer server along with human players or other bots.
    \item A bot can join an alternative server which implements the server-side component of the Minecraft network protocol. The development of the bot described in this paper uses the 3rd-party, open source \href{https://cuberite.org/}{Cuberite} server. Among other benefits, this server can be easily modified to record the game state that can be useful information to help improve the bot.
\end{enumerate}
\section{Assistant v0}

\begin{figure}
\centering \includegraphics[width=0.5\textwidth]{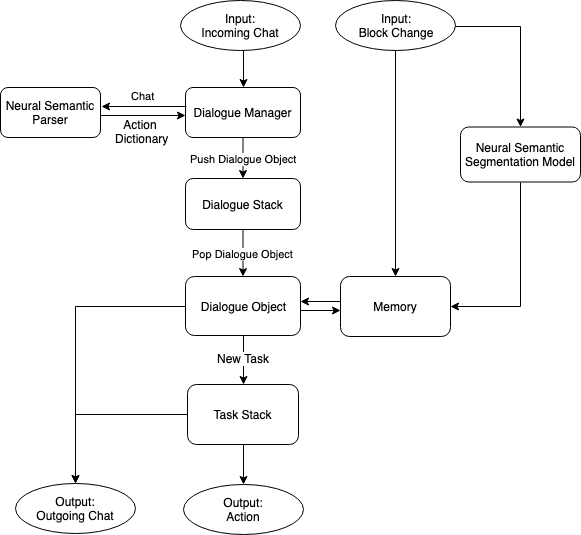}
\caption{A simplified block diagram demonstrating how the modular system reacts to incoming events (in-game chats and modifications to the block world)}
\label{fig:block_diagram}
\end{figure}

This section outlines our initial approach to building a Minecraft assistant, highlighting some of the major design decisions made:
%TODO: a little more detail on bullet points, and ensure link to discussion section is prominent
\begin{itemize}
    \item a modular architecture
    \item the use of high-level, hand-written composable actions called Tasks
    \item a pipelined approach to natural language understanding (NLU) involving a neural semantic parser
\end{itemize}
A simplified module-level diagram is shown in Figure \ref{fig:block_diagram}, and the code described here is available at: \href{https://github.com/facebookresearch/craftassist}{https://github.com/facebookresearch/craftassist}.  See Section \ref{sec:discussion} for a discussion of these decisions and our future plans to improve the bot.

Rather than directly modelling the action distribution as a function of the incoming chat text, our approach first parses the incoming text into a logical form we refer to as an \emph{action dictionary}, described later in section \ref{sec:action_dictionaries}. The action dictionary is then interpreted by a \emph{dialogue object} which queries the \emph{memory} module -- a symbolic representation of the bot's understanding of the world state -- to produce an action and/or a chat response to the user.

The bot responds to commands using a set of higher-level actions we refer to as Tasks, such as move to location $X$, or build a $Y$ at location $Z$. The Tasks act as abstractions of long sequences of low-level movement steps and individual block placements. The Tasks are executed in a stack (LIFO) order. The interpretation of an action dictionary by a dialogue object generally produces one or more Tasks, and the execution of the Task (e.g. performing the path-finding necessary to complete a \texttt{Move} command) is performed in a Task object in the bot's \emph{task stack}.

\subsection{Handling an Example Command}

Consider a situation where a human player tells the bot: \textbf{``go to the blue house''}. The Dialogue Manager first checks for illegal or profane words, then queries the semantic parser. The semantic parser takes the chat as input and produces the action dictionary shown in figure~\ref{fig:action_dict_go_to_the_house}. The dictionary indicates that the text is a command given by a human, that the high-level action requested is a \texttt{MOVE}, and that the destination of the \texttt{MOVE} is an object that is called a ``house" and is ``blue" in colour. More details on action dictionaries are provided in section~\ref{sec:action_dictionaries}. Based on the output of the semantic parser, the Dialogue Manager chooses the appropriate Dialogue Object to handle the chat, and pushes this Object to the Dialogue Stack.

In the current version of the bot, the semantic parser is a function of only text -- it is not aware of the objects present in the world. As shown in figure~\ref{fig:block_diagram}, it is the job of the Dialogue Object\footnote{The code implementing the dialogue object that would handle this scenario is in \href{https://github.com/facebookresearch/craftassist/blob/master/python/craftassist/dialogue_objects/interpreter.py}{interpreter.py}} to interpret the action dictionary in the context of the world state stored in the memory. In this case, the Dialogue Object would query the memory for objects tagged "blue" and "house", and  if present, create a \texttt{Move} Task whose target location is the actual $(x, y, z)$ coordinates of the blue house. More details on Tasks are in section~\ref{sec:tasks}

Once the Task is created and pushed onto the Task stack, it is the \texttt{Move} Task's responsibility, when called, to compare the bot's current location to the target location and produce a sequence of low-level step movements to reach the target.

\begin{figure}[ht]
\begin{Verbatim}[fontsize=\small]
Input: [0] "go to the blue house"
Output:
{
 "dialogue_type": "HUMAN_GIVE_COMMAND",
 "action": {
  "action_type": "MOVE",
  "location": {
   "location_type": "REFERENCE_OBJECT",
   "reference_object": {
    "has_colour": [0, [3, 3]],
    "has_name": [0, [4, 4]]
}}}}
\end{Verbatim}
\caption{An example input and output for the neural semantic parser. References to words in the input (e.g. "house") are written as spans of word indices, to allow generalization to words not present in the dictionary at train-time. For example, the word "house" is represented as the span beginning and ending with word 3, in sentence index 0.}
\label{fig:action_dict_go_to_the_house}
\end{figure}

\begin{figure}
\centering \includegraphics[width=0.3\textwidth]{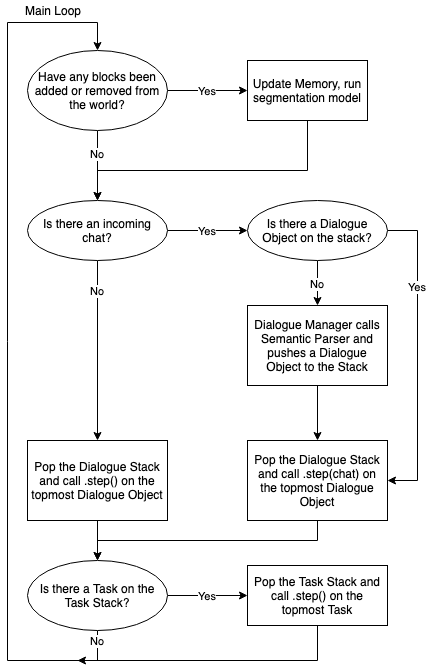}
\caption{A flowchart of the bot's main event loop. On every loop, the bot responds to incoming chat or block-change events if necessary, and makes progress on the topmost Task on its stack. Note that dialogue context (e.g. if the bot has asked a question and is awaiting a response from the user) is stored in a stack of Dialogue Objects. If this dialogue stack is not empty, the topmost Dialogue Object will handle an incoming chat.}
\label{fig:main_loop}
\end{figure}

A flowchart of the bot's main event loop is shown in figure~\ref{fig:main_loop}, and the implementation can be found in the \texttt{step} method in \href{https://github.com/facebookresearch/craftassist/blob/master/python/craftassist/craftassist_agent.py}{craftassist\_agent.py}.

%\section{Framework / Our Approach}
%\input{sec_framework.tex}
%\begin{itemize}
%    \item Semantic Parsing, action dictionaries, hand-crafted high-level Tasks
%    \item List and brief description of modules, block diagram
%    \item Flowchart / step-by-step walkthrough of two commands:
%    \begin{itemize}
%        \item A simple command ("destroy the blue house left of the barn")
%        \item A multi-turn dialogue
%    \end{itemize}
%\end{itemize}

\section{Modules}
This section provides a detailed documentation of each module of the system as implemented, at the time of this release.

\subsection{Task Stack}

\subsubsection{Task Primitives}

The following definitions are concepts used throughout the bot's Tasks and execution system:

\textbf{BlockId:  } A Minecraft building material (e.g. dirt, diamond, glass, or water), characterized by an 8-bit id and 4-bit metadata\footnote{See https://minecraft-ids.grahamedgecombe.com/ }

\textbf{Location:  } An absolute position  $(x, y, z)$ in the world

\textbf{Schematic:  } An object blueprint that can be copied into the world: a map of relative $(x, y, z) \mapsto \textrm{BlockId}$

\textbf{BlockObject:  } A real object that exists in the world: a set of absolute $(x, y, z)$

\textbf{Mob:  } A moving object in the world (e.g. cow, pig, sheep, etc.)

\subsubsection{Tasks} \label{sec:tasks}

A Task is an interruptible process with a clearly defined objective. A Task can be executed step by step, and must be resilient to long pauses between steps (to allow tasks to be paused and resumed if the user changes their priorities). A Task can also push other Tasks onto the stack, similar to the way that functions can call other functions in a standard programming language. For example, a Build may first require a Move if the bot is not close enough to place blocks at the desired location.

The following is a list of basic Tasks:

\textbf{Move(Location)} Move to a specific coordinate in the world. Implemented by an A* search which destroys and replaces blocks if necessary to reach a destination.

\textbf{Build(Schematic, Location)} Build a specific schematic into the world at a specified location.

\textbf{Destroy(BlockObject)} Destroy the specified BlockObject.

\textbf{Dig(Location, Size)} Dig a rectangular hole of a given Size at the specified Location.

\textbf{Fill(Location)} Fill the holes at the specified Location.

\textbf{Spawn(Mob, Location)} Spawn a Mob at a given Location. 

\textbf{Dance(Movement)} Perform a defined sequence of moves (e.g. move in a clockwise pattern around a coordinate)

There are also control flow actions which take other Tasks as arguments:

\textbf{Undo(Task)} This Task reverses the effects of a specified Task, or defaults to the last Task executed (e.g. destroy the blocks that resulted from a Build)

\textbf{Loop(StopCondition, Task)} This Task keeps executing the given Task until a StopCondition is met (e.g keep digging until you hit a bedrock block)

\subsection{Semantic Parser} \label{sec:semantic_parser}

The core of the bot's natural language understanding is performed by a neural semantic parser called the Text-to-Action-Dictionary (TTAD) model. This model receives an incoming chat / dialogue and parses it into an action dictionary that can be interpreted by the Dialogue Object.

A detailed report of this model is available at \cite{jernite2019craftassist}. The model is a modification of the approach in \cite{dong2016language}). We use bi-directional GRU encoder for encoding the sentences and multi-headed attention over the input sentence.

\subsubsection{Action Dictionaries}
\label{sec:action_dictionaries}

An action dictionary is an unambiguous logical form of the intent of a chat. An example of an action dictionary is shown in figure~\ref{fig:action_dict_go_to_the_house}. Every action dictionary is one of four dialogue types:

\begin{enumerate}
\item HUMAN\_GIVE\_COMMAND: The human is giving an instruction to the bot to perform a Task, e.g. to Move somewhere or Build something. An action dictionary of this type must have an \texttt{action} key that has a dictionary with an \texttt{action\_type} specifying the Task, along with further information detailing the information for the Task (e.g. ``schematic''  and ``location'' for a Build Task).
\item GET\_MEMORY: The human is asking a question or otherwise probing the bot's understanding of the environment.
\item PUT\_MEMORY: The human is providing information to the bot for future reference or providing feedback to the bot, e.g. assigning a name to an object ``that brown thing is a shed''.
\item NOOP: No action is required.
\end{enumerate}

There is a dialogue object associated with each dialogue type. For example, the \texttt{GetMemoryHandler} interprets a GET\_MEMORY action dictionary, querying the memory, and responding to the user with an answer to the question.

For HUMAN\_GIVE\_COMMAND action dictionaries, with few exceptions, there is a direct mapping from ``action\_type'' values to Task names in section \ref{sec:tasks}.

\subsection{Dialogue Manager \& Dialogue Stack}

The Dialogue Manager is the top-level handler for incoming chats. It performs the following :
\begin{enumerate}
    \item Checking the chat for obscenities or illegal words
    \item Calling the neural semantic parser to produce an action dictionary
    \item Routing the handling of the action dictionary to an appropriate Dialogue Object
    \item Storing (in the Dialogue Stack) persistent state and context to allow multi-turn dialogues
\end{enumerate}

The Dialogue Stack is to Dialogue Objects what the Task Stack is to Tasks. The execution of a Dialogue Object may require pushing another Dialogue Object onto the Stack. For example, the \texttt{Interpreter} Object, while handling a \texttt{Destroy} command and determining which object should be destroyed, may ask the user for clarification. This places a \texttt{ConfirmReferenceObject} object on the Stack, which in turn either pushes a \texttt{Say} object to ask the clarification question or \texttt{AwaitResponse} object (if the question has already been asked) to wait for the user's response. The Dialogue Manager will then first call the \texttt{Say} and then call the \texttt{AwaitResponse} object to help resolve the \texttt{Interpreter} object.

\subsection{Memory}
\label{sec:memory}

The data stored in the bot's memory includes the locations of BlockObjects and Mobs (animals), information about them (e.g. user-assigned names, colour etc), the historical and current state of the Task Stack, all the chats and relations between different memory objects. Memory data is queried by DialogueObjects when interpreting an action dictionary (e.g. to interpret the action dictionary in figure~\ref{fig:action_dict_go_to_the_house}, the memory is queried for the locations of block objects named ``house'' with colour ``blue'').

The memory module is implemented using an in-memory SQLite\footnote{https://www.sqlite.org/index.html} database. Relations and tags are stored in a single triple store. All memory objects (including triples themselves) can be referenced as the subject or object of a memory triple.

\paragraph{How are BlockObjects populated into Memory?} At this time, BlockObjects are defined as maximally connected components of unnatural blocks (i.e. ignoring blocks like grass and stone that are naturally found in the world, unless those blocks were placed by a human or bot). The bot periodically searches for BlockObjects in its vicinity and adds them to Memory.

\paragraph{How are tags populated into Memory?} At this time, tag triples of the form \texttt{(BlockObject\_id, "has\_tag", tag)} are inserted as the result of some PUT\_MEMORY actions, triggered when a user assigns a name or description to an object via chat or gives feedback (e.g. ``that object is a house'', ``that barn is tall'' or ``that was really cool"). Some relations (e.g. \texttt{has\_colour}, indicating BlockObject colours) are determined heuristically. Neural network perception modules may also populate tags into the memory.

\subsection{Perception}

The bot has access to two raw forms of visual sensory input:

\paragraph{2D block vision\footnote{The implementation of 2D block vision is found at \href{https://github.com/facebookresearch/craftassist/blob/master/client/src/agent.cpp\#L328}{agent.cpp\#L328}}} By default, this produces a 64x64 image where each ``pixel" contains the block type and distance to the block in the bot's line of sight. For example, instead of a pixel containing RGB colour information representing ``brown", the bot might see block-id 17, indicating ``Oak Wood".

\paragraph{3D block vision\footnote{The implementation of 3D block vision is found at \href{https://github.com/facebookresearch/craftassist/blob/master/client/src/agent.cpp\#L321}{agent.cpp\#L321}}} The bot has access to the underlying block map: the block type at any absolute position nearby. This information is not available to a human player interacting normally with the Minecraft game -- if it is important to compare a bot's sensorimotor capabilities to a human's (e.g. in playing an adversarial game against a human player), avoid the use of the \texttt{get\_blocks} function which implements this capability.

Other common perceptual capabilities are implemented using ML models or heuristics as appropriate:

\paragraph{Semantic segmentation} A 3d convolutional neural network processes each Block Object and outputs a tag for each voxel, indicating for example whether it is part of a wall, roof, or floor. The code for this model is in \href{https://github.com/facebookresearch/craftassist/tree/master/python/craftassist/vision/semantic\_segmentation}{python/craftassist/vision/semantic\_segmentation/}

\paragraph{Relative directions} Referring to objects based on their positions relative to other objects is performed heuristically based on a coordinate shift relative to the speaker's point of view. For example, referencing ``the barn left of the house" is handled by searching for the closest object called ``barn" that is to the speaker's left of the ``house".

\paragraph{Size and colour} Referring to objects based on their size or colour is handled heuristically. The colour of a Block Object is based on the colours of its most common block types. Adjectives referring to size (e.g. ``tiny" or ``huge") are heuristically mapped to ranges of block lengths.
%\begin{itemize}
%    \item Tasks + Task Stack
%    \item Semantic Parser (1-2 paragraphs + reference Arxiv paper)
%    \item Dialogue Manager, Dialogue Stack, Dialogue Objects, Interpreter(s)
%    \item Memory
%    \item Perception
%    \begin{itemize}
%        \item 2d/3d vision
%        \item instance segmentation
%    \end{itemize}
%\end{itemize}

\section{Data}
This section describes the datasets we are releasing with the framework. 
\subsection{The semantic parsing dataset}
We are releasing a semantic parsing dataset of English-language instructions and their associated ``action dictionaries'', used for human-bot interactions in Minecraft. 
This dataset was generated in different settings as described below:
\begin{itemize}
    \item \textbf{Generations}: Algorithmically generating action trees (logical forms over the grammar) with associated surface forms using templates. (The script for generating these is here: \href{https://github.com/facebookresearch/craftassist/blob/master/python/craftassist/ttad/generation_dialogues/generate_dialogue.py}{generate\_dialogue.py})
    \item \textbf{Rephrases}: We asked crowd workers to rephrase some of the produced instructions into commands in alternate, natural English that does not change the meaning of the sentence.
    \item \textbf{Prompts}: We presented crowd workers with a description of an assistant bot and asked them for examples of commands they'd give the bot.
    \item \textbf{Interactive}: We asked crowd workers to play creative mode Minecraft with our bot, and used the data from the in-game chat.
\end{itemize}

The dataset has four files, corresponding to the settings above:
\begin{enumerate}
    \item \textit{generated\_dialogues.json} : This file has 800000 dialogue - action dictionary pairs generated using our generation script. More can be generated using the script.
    \item \textit{rephrases.json}: This file has 25402 dialogue - action dictionary pairs. These are paraphrases of dialogues generated by our grammar.
    \item \textit{prompts.json}: This file contains 2513 dialogue - action dictionary pairs. These dialogues came from the prompts setting described above.
    \item \textit{humanbot.json}: This file contains 708 dialogue - action dictionary pairs. These dialogues came from the interactive setting above.
\end{enumerate}

The format of the data in each file is:
\begin{itemize}
    \item A dialogue is represented as a list of sentences, where each sentence is a sequence of words separated by spaces and tokenized using the spaCy tokenizer \cite{honnibal-johnson:2015:EMNLP}.
    \item Each json file is a list of dialogue - action dictionary pair, where ``action dictionary'' is a nested dictionary described in \ref{sec:action_dictionaries}
\end{itemize}

For more details on the dataset see: \cite{jernite2019craftassist}

\subsection{House dataset}
We used crowd sourcing to collect examples of humans building houses in Minecraft. Each user is asked to build a house on a fixed time budget (30 minutes), without any additional guidance or instructions. Every action of the user is recorded using the Cuberite server.

The data collection was performed in Minecraft's creative mode, where the user is given unlimited resources, has access to all material block types and can freely move in the game world. The action space of the environment is straight-forward: moving in x-y-z dimensions, choosing a block type, and placing or breaking a block.

There are hundreds of different block types someone could use to build a house, including different kinds of wood, stone, dirt, sand, glass, metal, ice, to list a few. An empty voxel is considered as a special block type ``air'' (block id=0).

We record sequences of atomic building actions for each user at each step using the following format:
\begin{center}
\begin{BVerbatim}
[t, userid, [x, y, z], 
    [block-id, meta-id], "P"/"B"]
\end{BVerbatim}
\end{center}
where the time-stamp $t$ is in monotonically increasing order;  $[x_t, y_t, z_t]$ is the absolute coordinate with respect to the world origin in Minecraft; ``P'' and ``B'' refers to placing a new block and breaking (destroying) an existing block; each house is built by a single player in our data collection process with a unique user-id. 

There are 2586 houses in total. Details of this work is under submission.

\subsection{Instance segmentation data}
For a subset of the houses collected in the house dataset described above, we asked crowd workers to add semantic segmentation labels for sub-components of the house.
The format of the data is explained below.
There are two files:
\begin{itemize}
    \item \textbf{training\_data.pkl} : This file contains data we used for training our 3D semantic segmentation model.
    \item \textbf{validation\_data.pkl}: This file contains data used as validation set for the model.
\end{itemize}

Each pickle file has a list of :
\begin{center}
\begin{BVerbatim}
[schematic, annotated_schematic, 
      annotation_list, house_name]
\end{BVerbatim}
\end{center}
where:
\begin{itemize}
    \item \textit{schematic}: The 3-d numpy array representing the house, where each element in the array is the block\_id of the block at that coordinate.
    \item \textit{annotated\_schematic}: The 3-d numpy array representing the house, where each element in the array is the id of the semantic annotation that the coordinate/block belongs to (1-indexed annotation\_list).
    \item \textit{annotation\_list}: List of semantic segmentation for the house.
    \item \textit{house\_name}: Name of the house.
\end{itemize}
There are 2050 houses in total and 1038 distinct labels of subcomponents.

The datasets described above can be downloaded following the instructions \href{https://github.com/facebookresearch/craftassist#datasets}{here}

\section{Related Work}
A number of projects have been initiated to study Minecraft agents or to build frameworks to make learning in Minecraft possible.  The most well known framework is Microsoft's MALMO project \cite{johnson2016malmo}.
The majority of work using MALMO consider reinforcement learned agents to achieve certain goals e.g \cite{shu2017hierarchical,udagawa2016fighting,alaniz2018deep,oh2016control,tessler2017deep}.   Recently the MineRL project \cite{MineRL} builds on top of MALMO with playthrough data and specific challenges.

Our initial bot has a neural semantic parser  \cite{dong2016language,jia2016data, zhong2017seq2sql} as its core NLU component.   We also release the data used to train the semantic parser.  There have been a number of datasets of natural language paired with logical forms to evaluate semantic parsing approaches, e.g. \cite{price1990evaluation, tang2001using, cai2013large, wang2015building, zhong2017seq2sql}.   Recently \cite{chevalier2018babyai} described a gridworld with navigation instructions generated via a grammar. Our bot also needs to update its understanding of an initial instruction during several turns of dialogue with the user, which is reminiscent of the setting of \cite{DBLP:conf/iclr/BordesBW17}.

In addition to mapping natural language to logical forms, our dataset connects both of these to a dynamic environment.    In \cite{tellex2011understanding, matuszek2013learning} semantic parsing has been used for interpreting natural language commands for robots. %, matuszek2013learning}.  
In our setup, the ``robot'' is embodied in the Minecraft game instead of in the physical world.
Semantic parsing in a voxel-world recalls \cite{wang2017naturalizing}, where the authors describe a method for building up a programming language from a small core via interactions with players.   Our bot's NLU pipeline is perhaps most similar to the one proposed in \cite{kollar2018alexa}, which builds a grammar for the Alexa virtual personal assistant.

A task relevant to interactive bots is that of Visual Question Answering (VQA) ~\cite{antol2015vqa, krishna2017visual, geman2015visual} in which a question is asked about an image and an answer is provided by the system. Most papers address this task using real images, but synthetic images have also been used \cite{CLEVR, andreas2016neural}. The VQA task has been extended to visual dialogues~\cite{das2017visual} and videos~\cite{tapaswi2016movieqa}. Recently, the tasks of VQA and navigation have been combined using 3D environments~\cite{gordon2018iqa, das2018embodied, kolve2017ai2, anderson2018vision} to explore bots that must navigate to certain locations before a question can be answered, e.g., ``How many chairs are in the kitchen?'' Similar to our framework, these papers use synthetic environments for exploration. However, these can be expanded to use those generated from real environments~\cite{savva2019habitat}. Instead of the goal being the answering of a question, other tasks can be explored. For instance, the task of guiding navigation in New York City using dialogue~\cite{de2018talk}, or accomplishing tasks such as pushing or opening specific objects~\cite{kolve2017ai2}.

\section{Discussion}
\label{sec:discussion}

%\begin{itemize}
%    \item Design decision pros/cons
%    \item Semantic Parsing vs. language->actions
%    \item Semantic Parser unaware of world state
%    \item Symbolic Memory?

% - current limitations and what could be done in future -- but not flag planting!
% - call to arms to edit, use pieces/all of what we've done, make different decisions
% - 
%\end{itemize}

In this work we have described the design of a bot and associated data that we hope can be used as a starting point and baseline for research in learning from interaction in Minecraft.   In this section, 
we discuss some major design decisions that were made for this bot, and contrast against other possible choices.   We further discuss ways in which the bot can be improved.

%We wish to be clear that we do not expect other researchers to necessarily make the same design choices.  However, we expect that for many research programs, pieces of (if not the whole) framework will be useful.

\subsection{Semantic Parsing}

Rather than learning a mapping directly from (language, state) to an action or sequence of actions, the bot described in this paper first parses language into a program over high level tasks, called action dictionaries (see section~\ref{sec:action_dictionaries}).   The execution of the program is scripted, rather than learned.

This arrangement has several advantages:
\begin{enumerate}
    \item Determining a sequence of actions, given a well-specified intent, is usually simple in Minecraft. For example, moving to a known but faraway object might require hundreds of steps, but it is simple to use a path-finding algorithm such as A* search to find the sequence of actions to actually execute the move. 
    %than to train an ML model to produce this long sequence.
    \item Training data for a semantic parsing model is easier to collect, compared to language-action pairs that would necessitate recording the actions of a human player.
    \item If it was desired to learn the low-level actions needed to complete a task, approaches such as reinforcement learning could be employed that use the completion of the task in the action dictionary as a reward without having to address the ambiguities of tasks specified through language.
    \item It may be possible to transfer the natural language understanding capabilities of the bot to another similar domain by re-implementing the interpretation and execution of action dictionaries, without needing to retrain the semantic parser.
\end{enumerate}
On the other hand, 
\begin{enumerate}
    \item The space of objectives that can be completed by the bot is limited by the specification of the action dictionaries. Adding a new capability to the bot usually requires adding a new structure to the action dictionary spec, adding code to the relevant Dialogue Object to handle it, and updating the semantic parsing dataset.
    \item A more end-to-end model with a simpler action space might only require the collection of more data and might generalize better.
    \item The use of a pipelined approach (as described in this paper) introduces the possibility for compounding errors.
    % \item In our current implementation, the action dictionaries must be designed so that they can be inferred from text {\it without} access to the world state.  More generally, the representation of the text of commands or dialogues is not tied to the representation of the world-state or how actions affect the world-state. \lz{This is more a point for how that bot can be improved, rather than a problem with using an action dictionary.}
\end{enumerate}

%In our view, training a model to output low-level actions is counterproductive in the Minecraft domain, and we expect later versions of the bot to continue using a high level interface.  

% Nevertheless, 
There is a huge space of possible interfaces into the  high-level actions we have proposed (and many other interesting constructions of high level actions).  In particular, we plan to remove the strict separation between the parser and the world state in our bot.

\subsection{Symbolic Memory}

As described in section~\ref{sec:memory}, the bot's memory is implemented using a (discrete, symbolic) relational database.  The major advantages of this (compared to an end-to-end machine-learned model that operates on raw sensory inputs) are:
\begin{enumerate}
    \item Easier to convert semantic parses into fully specified tasks that can query and write to the database.
    \item Debugging the bot's current understanding of the world is easier.
    \item Integrating outside information, e.g. crowd-sourced building schematics, is more straightforward: doing so requires pre-loading rows into a database table, rather than re-training a generative model.
    \item Reliable symbolic manipulations, especially lookups by keyword.
\end{enumerate}

On the other hand, such a memory can be brittle and limited.  Even within the space of ``discrete'' memories, there are more flexible formats, e.g. raw text; and there have been recent successes using such memories, for example works using the Squad dataset \cite{rajpurkar2016squad}.   We hope our platform will be useful for studying other symbolic memory architectures as well as continuous, learned approaches, and things in between.

\subsection{Modularity for ML research}
The bot's architecture is modular, and currently many of the modules are not learned.   Many machine learning researchers consider the sort of tasks that pipelining makes simpler to be tools for evaluating more general learning methodologies.   Such a researcher might advocate more end-to-end (or otherwise less ``engineered'') approaches because the goal is not necessarily to build something that works well on the tasks that the engineered approach can succeed in, but rather to build something that can scale beyond those tasks.

 We have chosen this approach in part because it allows us to more easily build an interesting initial assistant from which we can iterate; and in particular allows data collection and creation.  We do believe that modular systems are more generally interesting, especially in the setting of competency across a large number of relatively easier tasks.    Perhaps most interesting to us are approaches that allow modular components with clearly defined interfaces, and heterogeneous training based on what data is available.  We hope to explore these with further iterations of our bot.

Despite our own pipelined approach, we consider research on more end-to-end approaches worthwhile and interesting.  Even for researchers primarily interested in these, the pipelined approach still has value beyond serving as a baseline:  as discussed above, it allows generating large amounts of training data for end-to-end methods.

Finally, we note that from an engineering standpoint, modularity has clear benefits.  In particular, it allows many researchers to contribute components to the greater whole in parallel.  As discussed above, the bot presented here is meant to be a jumping off point, not a final product.  We hope that the community will find the framework useful and join us in building an assistant that can flexibly learn from interaction with people. 

\begin{comment}
\subsection{Community engagement}

 %Our own perspective is that we should take the specific challenges of building a Minecraft assistant at face value.
 
 The ability to accomplish a task as ambitious as CraftAssist would benefit from expertise in multiple sub-fields, such as NLP and computer vision, in addition to the broader machine learning community. To help empower the research community in exploring bots in CraftAssist, we are open-sourcing the software-framework for our initial bot. 
 
 The modular design of our bot has clear benefits.  In particular, it allows many researchers to contribute components to the greater whole in parallel.  As discussed above, the bot presented here is meant to be a jumping off point, not a final product.  
 
 In addition to the code, we are releasing a diverse set of data to help in the training bots. In the future, we plan to continue to release data as it is collected. We hope that the community will find the framework useful and join us in building an assistant that can flexibly learn from interaction with people. 
 \end{comment}

\section{Conclusion}
We have described a platform for studying situated natural language understanding in Minecraft.  The platform consists of code that implements infrastructure for allowing bots and people to play together, tools for labeling data, and a baseline assistant.   In addition to the code, we are releasing a diverse set of data we used for building the assistant.  This includes 2586 houses built in game, and the actions used in building them, instance segmentations of those houses, and templates and rephrases of templates for training a semantic parser. In the future, we plan to continue to release data as it is collected. We hope that the community will find the framework useful and join us in building an assistant that can learn a broad range of tasks from interaction with people. 

\bibliography{main}
\bibliographystyle{icml2019}

\end{document}